\documentclass{article}

\usepackage[final]{neurips_2022}

\usepackage{color}
\usepackage[dvipsnames]{xcolor}

\newcommand\przemek[1]{ #1 }

\def\our{Points2NeRF}

\usepackage{wrapfig}

\usepackage[utf8]{inputenc} % allow utf-8 input
\usepackage[T1]{fontenc}    % use 8-bit T1 fonts
\usepackage{hyperref}       % hyperlinks
\usepackage{url}            % simple URL typesetting
\usepackage{booktabs}       % professional-quality tables
\usepackage{amsfonts}       % blackboard math symbols
\usepackage{nicefrac}       % compact symbols for 1/2, etc.
\usepackage{microtype}      % microtypography
\usepackage{xcolor}         % colors
\usepackage{multirow}       %multirows in table
\usepackage{graphicx}
\usepackage{bookmark}

\def\N{\mathbb{N}}
\def\Z{\mathbb{Z}}
\def\R{\mathbb{R}}

\def\1{\mathbbm{1}}

\def\X{\mathcal{X}}
\def\Z{\mathcal{Z}}
\def\E{\mathcal{E}}
\def\D{\mathcal{D}}

\title{ \our{}: Generating Neural Radiance Fields from 3D point cloud}

\author{%
  Dominik Zimny \\
  Jagiellonian University, Kraków, Poland\\
  \And
  Joanna Waczyńska \\
  Jagiellonian University, Kraków, Poland\\
  \And
Tomasz Trzciński \\
  Warsaw University of Technology, Warsaw, Poland, \\
IDEAS NCBR\\
\And
Przemysław Spurek \\
  Jagiellonian University, Kraków, Poland\\
IDEAS NCBR\\
}

\begin{document}

\maketitle

\begin{abstract}
Neural Radiance Fields (NeRFs) offers a state-of-the-art quality in synthesizing novel views of complex 3D scenes from a small subset of base images. For NeRFs to perform optimally, the registration of base images has to follow certain assumptions, including maintaining a constant distance between the camera and the object. We can address this limitation by training NeRFs with 3D point clouds instead of images, yet a straightforward substitution is impossible due to the sparsity of 3D clouds in the under-sampled regions, which leads to incomplete reconstruction output by NeRFs. To solve this problem, here we propose an auto-encoder-based architecture that leverages a hypernetwork paradigm to transfer 3D points with the associated color values through a lower-dimensional latent space and generate weights of NeRF model. This way, we can accommodate the sparsity of 3D point clouds and fully exploit the potential of point cloud data. As a side benefit, our method offers an implicit way of representing 3D scenes and objects that can be employed to condition NeRFs and hence generalize the models beyond objects seen during training. The empirical evaluation confirms the advantages of our method over conventional NeRFs and proves its superiority in practical applications.
\end{abstract}

%\linenumbers

%% main text

\section{Introduction}

Neural Radiance Fields (NeRFs)~\cite{mildenhall2020nerf} enables synthesizing novel views of complex scenes from a few 2D images with known camera positions. Based on the relations between those base images and computer graphics principles, such as ray tracing, this neural network model can render high-quality scene images from previously unseen viewpoints. Although in recent years, much effort was invested in improving the quality of the resulting views~\cite{kosiorek2021nerf} and the controllability of NeRFs~\cite{kania2021conerf}, the robustness of those methods against various data registration challenges remain largely unchartered research area. For instance, to render high-quality views, the base images must be captured from a distance similar to the captured object, and the corresponding camera positions must be approximately known. 
\przemek{ Such a problem is partially solved in Mip-nerf~\cite{barron2021mip}, where authors reduce aliasing problems during rendering. But steel, we train on images taken from a fixed distance.} 
These practical constraints limit the applicability of NeRFs across various applications, such as in robots deployed within noisy industrial environments.

\begin{figure*}[h!]

\quad 
\qquad  Point Cloud \qquad \qquad \qquad \qquad \qquad KNN-NeRF \qquad \qquad \qquad \qquad \qquad \quad \our{} \\ 
\begin{centering}
	\includegraphics[width=0.9\linewidth]{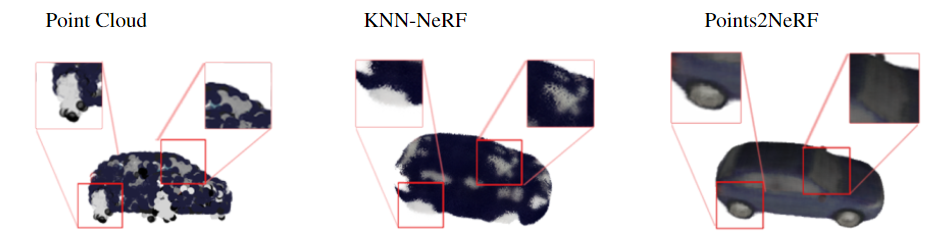} 
\end{centering}		
 	% \vspace{-0.5cm}

\caption{Conversely to other approaches, our {\our{}} method takes a unstructured 3D point cloud with colors as an input instead of 2D images. A straightforward substitution of the input data in the existing architectures is not feasible due to the sparsity of point cloud representation ({\bf left} image), which yields artifacts in reconstruction (baseline KNN-NeRF result in the {\bf middle}). By extending an auto-encoder architecture with the hypernetwork paradigm, our {\our{}} architecture can build an internal representation of complete 3D objects, which enhances the rendering quality of NeRFs ({\bf right} image).}
\label{fig:teser1} 
\end{figure*}

\przemek{We can address these limitations by using increasingly popular 3D capturing devices, such as LIDARs and depth cameras, to produce 3D point clouds and feed those point clouds to NeRFs. A straightforward substitution of 2D images with unstructured 3D point clouds in classic NeRF produces novel views with lower quality than image-based NeRF. This is mainly because of the sparsity of point cloud data in the under-sampled parts of objects, {\it, e.g.} willows and windows of a car presented in Fig.~\ref{fig:teser1}. As a result, the baseline solution, dubbed KNN-NeRF and described in detail in Sec.~\ref{sec:knn}, does not render sharp images.}

To solve this problem, we propose an auto-encoder-based architecture that transfers 3D point clouds through a low-dimensional latent space and outputs weights of a NeRF model. Our {\our{}}\footnote{We make our implementation available at \url{https://github.com/gmum/points2nerf}} approach embodies hypernetwork paradigm as we take a 3D point cloud with the associated color values as an input and return the parameters of a target NeRF neural network. Since the implicit representation of the captured objects is built within the latent space, missing data points in under-sampled regions do not prohibit the resulting target network from synthesizing high-quality views. Furthermore, during the training, our model can process multiple classes of point cloud objects, thus yielding a solution that is much more robust. The NeRF network parameters can be interpreted as a continuous parametrization of a 3D object space. Thus, this formulation allows for conditioning NeRFs and generalizing our solution beyond 3D object classes seen during training.

To summarize, the contributions of our work are the following:
\begin{itemize}
    \item We propose a new method dubbed~\our{} which adapts a hypernetwork framework to the NeRF architecture and hence allows to production of Radiance Fields from a 3D point cloud.
    \item Our approach enables conditioning NeRFs, which, in turn, allows us to generalize the model beyond 3D objects seen in training.
    \item Lastly, our method offers a generative model that can continuously represent 3D objects as NeRF parameters, enabling interpolation within the object space.
\end{itemize}

\begin{figure*}[t!]
\includegraphics[width=0.9\linewidth]{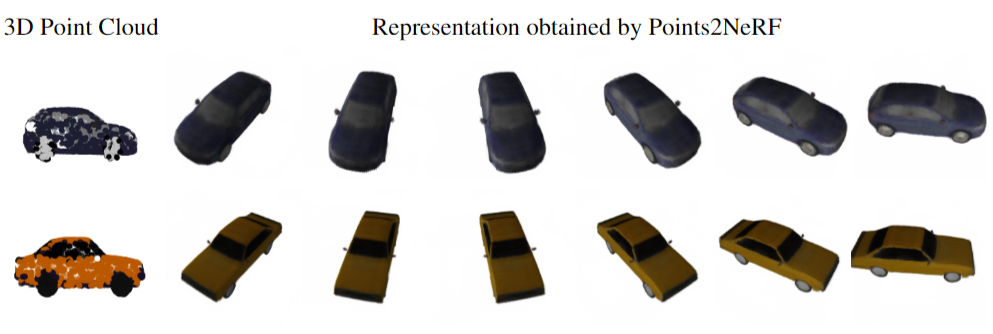} 
\caption{ Our \our{} approach takes a 3D point cloud with the associated color values and returns the weights of a NeRF network that reconstructs 3D objects with high fidelity and coherent coloring.}
\label{fig:teser2} 
\end{figure*}

\przemek{The paper is arranged as follows. In the next section, we present the related works. In 
 Section~\ref{sec:knn}, we discuss  all problems with training NeRF only on 3D point clouds without images. In Section~\ref{sec_our} we describe our model \our{}. In the last section, we present experiments.}

%%%%%%%%%%%%%%%%%%%%%%%%%%%%%%%%%%%%%%%%%%%
\section{Related works}
\label{gen_inst}
%%%%%%%%%%%%%%%%%%%%%%%%%%%%%%%%%%%%%%%%%%%

3D objects can be represented by using various techniques, including voxel grids~\cite{choy20163d}, octrees \cite{hane2017hierarchical}, multi-view images \cite{arsalan2017synthesizing,LIU2022108774}, point clouds \cite{achlioptas2018learning,shu2022wasserstein,yang2022continuous}, geometry
images \cite{sinha2016deep}, deformable meshes \cite{girdhar2016learning},
and part-based structural graphs \cite{li2017grass}.

The above representations are discreet, a substantial limitation in real-life applications. Alternatively, we can represent 3D objects as a continuous function~\cite {dupont2021generative}. In practice implicit occupancy~\cite{chen2019learning,mescheder2019occupancy,peng2020convolutional}, distance field~\cite{michalkiewicz2019implicit,park2019deepsdf} and surface parametrization~\cite{yang2019pointflow,spurek2020hypernetwork,spurek2021general,cai2020learning} models use a neural network to parameterize a 3D object. In such a case, we do not have a fixed number of voxels, points, or vertices, but we represent shapes as a continuous function. 

These models are limited by their requirement of access to ground truth 3D geometry. Recently work relaxed this requirement of ground truth 3D shapes using only 2D images. In \cite{niemeyer2020differentiable}, authors present 3D occupancy fields. The numerical method is used to find the surface intersection for each ray. In \cite{sitzmann2019scene}, propose a neural network that produces feature vector and RGB color at each continuous 3D coordinate, and propose a differentiable rendering function consisting of a recurrent neural network.

The above models are limited to simple shapes with low geometric complexity, resulting in over-smoothed renderings. To solve such a problem, NeRF model was proposed. NeRF represents a static scene as a continuous 5D function that outputs the radiance emitted in each direction and a density at each point which acts like a differential opacity controlling how much radiance is accumulated by a ray passing through the point.

The NeRF method is a state-of-the-art solution for representing 3D objects. However, the model has many different generalizations for static and dynamic scenes.
\przemek{ Neural Sparse Voxel Fields \cite{liu2020neural} is a modification of NeRF where we use trainable voxel representation of 3D objects instead of a relatively large neural network. Such an approach allows the training of NeRF to be faster. In Plenoxels~\cite{fridovich2022plenoxels}, authors propose to train only voxels in neighborhood nonzero colors. This method is faster since we do not need to train an entire voxel grid.
KiloNeRF \cite{reiser2021kilonerf} solves a similar problem, but we use many small NeRF which render small parts of the object.    }
\przemek{ In Mip-NeRF~\cite{barron2021mip} authors propose a method that reduces the aliasing problem when the camera distance to render object is changing. }

Information from point clouds was used in NeRF across  different applications. \przemek{In \cite{xu2022point} authors present a novel neural scene representation Point-NeRF that models a volumetric radiance field with a neural point cloud.
Point-NeRF, similar to Points2NeRF, uses a 3D point cloud. But in a completely different fashion. In Point-NeRF, a 3D point cloud is used to help the model render new views correctly. In practice, we train Point-NeRF on a single object using 2D images and a 3D point cloud. Point clouds are generated from a depth camera or arbitrary given.  
In \our{}, we use 3D point clouds and 2D images in training. But point clouds are input to the autoencoder, and 2D images go to the resulting NeRF. After training, we use only 3D point clouds to input the autoencoder. In inference time, we use only the 3D point cloud and produce NeRF, which can generate 2D images.
} In NeuS~\cite{wang2021neus}, authors propose to represent a surface as the zero-level set of a signed distance function (SDF) and develop a new volume rendering method to train a neural SDF representation. In consequence, we obtain a novel neural surface reconstruction method. In \cite{ost2021neural}, authors introduce Neural Point Light Fields that represent scenes implicitly with a light field living on a sparse point cloud. We use point clouds to produce NeRF representation in our work. 

Most such models are trained on a single scene. 
NeRF-VAE~\cite{kosiorek2021nerf} is a 3D scene generative model. In contrast to NeRF, such a model considers shared structure across scenes.
Unfortunately, the model was trained only on simple scenes containing geometric figures. In our paper, we present \our{}, which is trained on an extensive data set and can transform a 3D point cloud to NeRF representation.

%%%%%%%%%%%%%%%%%%%%%%%%%%%%%%%%%%%%%%%%%%%%%
\section{Training NeRF based on 3D point cloud}\label{sec:knn}
%%%%%%%%%%%%%%%%%%%%%%%%%%%%%%%%%%%%%%%%%%%%%

NeRFs~\cite{mildenhall2020nerf} represent a scene using a fully-connected architecture. As the input, NeRF takes a 5D coordinate (spatial location $ {\bf x} = (x, y, z)$ and viewing direction ${\bf d} =  (\theta, \psi)$) and it outputs an emitted color ${\bf c} = (r, g, b)$ and volume density $\sigma$.

%%%%%%%%%%%%%%%%%%%%%%%%%%%%%%%%%%%%%%%%%%%
\subsection{NeRFs trained on images.} 
%%%%%%%%%%%%%%%%%%%%%%%%%%%%%%%%%%%%%%%%%%%

A vanilla NeRF uses a set of images for training. In such a scenario, we produce many rays passing through the image and a 3D object represented by a neural network. 
NeRF approximates this 3D object with MLP network:
$$
F_{\Theta} : ({\bf x}, {\bf d}) \to ( {\bf c} , \sigma),
$$
and optimizes its weights $\Theta$ to map each input 5D coordinate to its corresponding volume density and directional emitted color. 

The loss of NeRF is inspired by classical volume rendering \cite{kajiya1984ray}.  We render the color of all rays passing through the scene. The volume density $\sigma( {\bf x} )$ can be interpreted as the differential probability of a ray. The expected color
$C({\bf r})$ of camera ray ${\bf r}(t) = {\bf o} + t {\bf d} $ (where  ${\bf o}$ is ray origin and ${\bf d}$ is direction) can be computed with an integral.

In practice, this continuous integral is numerically estimated using  a quadrature. We use a stratified sampling approach where we partition our ray $[t_n, t_f ]$ into $N$ evenly-spaced bins and then draw one sample uniformly at random from within each bin:
$$
t_i \sim \mathcal{U}[t_n+\frac{i-1}{N}(t_f-t_n), t_n+\frac{i}{N}(t_f-t_n) ].
$$
We use these samples to estimate $C({\bf r})$ with the quadrature rule
discussed in the volume rendering review by Max~\cite{max1995optical}:
\begin{equation}
\hat C({\bf r}) = \sum_{i=1}^{N} T_i (1-\exp(-\sigma_i \delta_i)) {\bf c}_i, \mbox{ where } T(t)=\exp \left(-\sum_{j=1}^{i-1} \sigma_i \delta_i \right).
\end{equation}
where $\delta_i = t_{i+1} - t_i$ is the distance between adjacent samples. This function
for calculating $\hat C({\bf r})$ from the set of $(c_i , \sigma_i)$ values trivially differentiable.

We then use the volume rendering procedure  to render the color of each ray
from both sets of samples. Our loss is simply the total squared error between
the rendered and true pixel colors 
\begin{equation}
    \mathcal{L} = \sum _{{\bf r} \in R} \| \hat C({\bf r}) - C({\bf r}) \|_2^2
    \label{eq:cost_general}
\end{equation}

where $R$ is the set of rays in each batch, and $C({\bf r})$, $\hat C ({\bf r})$ are the ground truth and predicted RGB colors for ray {\bf r } respectively. Contrary to the baseline NeRF~\cite{mildenhall2020nerf}, we use only a single architecture.

%%%%%%%%%%%%%%%%%%%%%%%%%%%%%%%%%%%%%%%%%%%%%%%%%%%%%%%%%%%%%%
\subsection{NeRFs trained on point clouds} 
%%%%%%%%%%%%%%%%%%%%%%%%%%%%%%%%%%%%%%%%%%%%%%%%%%%%%%%%%%%%%%

Such an approach can be easily modified to train NeRFs using a 3D point cloud. We take an input point cloud $X \subset \R^6$ where the first three elements encode the position while the last three encode an RGB color.

We can use only rays which cross points from \przemek{an unstructured 3D point cloud}. Let $ {\bf x} = (x,y,z,r,g,b) \in X $ be a point from our point cloud with color. By ${\bf {\bf x}_p} = (x,y,z)$ we denote the coordinates of the point and by ${\bf {\bf x}_c} = (r,g,b)$ we denote color of the point ${\bf x} = ({\bf x}_p, {\bf x}_c)$. The ray going through point ${\bf x}_p$ from origin ${\bf o}$ is defined by 
$$
{\bf r}_{ {\bf x}_p }(t) = {\bf o} + t\frac{ { {\bf x}_p }-  {\bf o} }{ \| { {\bf x}_p }-{\bf o} \|}.
% \mbox{ where } {\bf d}_{\bf p} = \frac{ {\bf p}-  {\bf o} }{ \| {\bf p}-{\bf o} \|}. 
$$
In such a case as a ground true color, we use the color of the point laid on the ray instead of the pixel from the image
$C({\bf r}_{\bf {\bf x}_p}) = {\bf x}_c$.
Our loss is simply the total squared error between
the rendered and true point colors 
\begin{equation}
    \mathcal{L} = \sum _{  ({\bf x}_p, {\bf x}_c)  \in X} \| \hat C({\bf r}_{{\bf x}_p}) - {\bf x}_c ) \|_2^2.
    % = \sum _{  (x,y,z,r,g,b)  \in X} \| \hat C({\bf r}_{(x,y,z)}) - (r,g,b ) \|_2^2
    \label{eq:cost_general_point}
\end{equation}

\przemek{Unfortunately, such an approach cannot reconstruct the correct image since we pass rays only thru a limited number of pixels on the rendered images. 
In practice, we use only rays going through the existing 3D points. The model is not able to extrapolate renders on unseen parts of the rendered images.  In particular, the model does not see the object borders and cannot reconstruct them sharply.}

%%%%%%%%%%%%%%%%%%%%%%%%%%%%%%%%%%%%%%%%
\subsection{KNN approach to training NeRF on point cloud}
%%%%%%%%%%%%%%%%%%%%%%%%%%%%%%%%%%%%%%%%

To solve this problem, we can produce many different rays not restricted to the points from training data. In such a scenario, we do not cross any points from the point cloud. Therefore, as a ground truth color, we can use a color of the point which is closest to our ray. It is not trivial to find such an element since our ray goes through a 3D object and crosses the front and back of the object. So the point must fulfill two main constraints. It must be as close as possible to points on the ray and the origin of the ray. 

We use the K-nearest neighbor (KNN) algorithm to solve the problem. Let us consider point cloud $X \subset \R^6$ where the first three elements encode the position while the last three encode an RGB color and ray ${\bf r}(t) = {\bf o} + t {\bf d} $ (where  ${\bf o}$ is ray origin and ${\bf d}$ is direction).
In the first step, we find $k \in \N$ closest elements to the origin $KNN_{\bf o}^k(X)$. Thanks to this procedure, we obtain $KNN_{\bf o}^k(X)$, which contains elements from the surface of $X$ situated on the front of the object (according to origin ${\bf o}$). In practice,  $k$ is a hyperparameter. 
In the second step, we look for the closest element to the ray ${\bf r}$:
$$
  C(X, {\bf r}) = \left\{ {\bf \bar x} \in KNN_{\bf o}^k(X) : d( {\bf \bar x_p}, {\bf r} ) \leq  d( {\bf x_p}, {\bf r} ) \mbox{ for } {\bf x} \in  KNN_{\bf o}^k(X) \right\}
$$
where $d(\cdot, \cdot)$ is a distance from a point to a line.

When we find the closest element to the ray, we can use its color as a ground truth
\begin{equation}
    \mathcal{L} = \sum _{{\bf r} \in R} \| \hat C({\bf r}) - C_c(X, \bf r) \|_2^2
    \label{eq:cost_general_knn}
\end{equation}

where $ C_c(X, \bf r)$ is the color of closest point from $X$ to ray ${\bf r}$ and $\hat C(r)$ is the  predicted RGB colors for ray ${\bf r}$. 

This solution, which we dub KNN-NeRF, can be used as a baseline for our model. Unfortunately, KNN-NeRFs cannot reconstruct correct shapes since point clouds in training are highly sparse. \przemek{ Even in original NeRF paper~\cite{mildenhall2020nerf} authors highlight that NeRFs perform poorly at representing high-frequency variation in color and geometry. This is consistent with recent work by~\cite{rahaman2019spectral}, which shows that deep networks are biased towards learning lower frequency functions. }
To solve this limitation, we propose an auto-encoder-based architecture in this work, which transfers the 3D point cloud into NeRF, mitigating the problem with sparse input data.

%%%%%%%%%%%%%%%%%%%%%%%%%%%%%%%%%%%%%%%%%%%%%
\section{\our{}: generating Neural Radiance Fields
from 3D point cloud} \label{sec_our}
%%%%%%%%%%%%%%%%%%%%%%%%%%%%%%%%%%%%%%%%%%%%%

In this section, we present our \our{} model for building NeRF representations of 3D point clouds. To that end, we leverage three main components described below: the auto-encoder architecture, the NeRF representation of a 3D static scene, and the hypernetwork training paradigm. 
The central intuition behind our approach is the construction of an auto-encoder, which takes as an input 3D point cloud and generates the weight of the target network -- NeRF. For practical NeRF training, our model requires a set of 2D images on top of the 3D point clouds captured by 3D registration devices. 

%%%%%%%%%%%%%%%%%%%%%%%%%%%%%%%%%%%%
\subsection{Hypernetwork}
%%%%%%%%%%%%%%%%%%%%%%%%%%%%%%%%%%%%

Hypernetworks, introduced in~\cite{ha2016hypernetworks}, are defined as neural models that generate weights for a separate target network solving a specific task. The authors aim to reduce the number of trainable parameters by designing a Hypernetwork with smaller parameters. Making an analogy between Hypernetworks and generative models, the authors of \cite{sheikh2017stochastic} use this mechanism to generate a diverse set of target networks approximating the same function. 

In the context of 3D objects, various methods use hypernetwork to produce a continuous representation of objects~\cite{spurek2021hyperpocket,spurek2020hypernetwork,spurek2021general}. HyperCloud~\cite{spurek2020hypernetwork} represents a 3D point cloud as a classical MLP, while in \cite{spurek2021general}, it is represented by a Continuous Normalizing Flow~\cite{grathwohl2018ffjord}. 
 
%%%%%%%%%%%%%%%%%%%%%%%%%%%%%%%%%%%%
\subsection{Auto-encoders for 3D Point Clouds}
%%%%%%%%%%%%%%%%%%%%%%%%%%%%%%%%%%%%

Our approach uses a hypernetwork paradigm to aggregate information from 3D point cloud representation and produces the weights of a NeRF architecture. Moreover, such a solution allows us to create a high-resolution model of the 3D point cloud.

In \our{} hypernetwork is an auto-encoder type architecture for the 3D point cloud. Let 
$\X = \{X_i\}_{i=1,\ldots,n} = \{(x_i,y_i,z_i,r_i,g_i,b_i)\}_{i=1,\ldots,n}$  be a given dataset containing point cloud with colors. The first three
elements encode the position, while the last three encode  RGB color. The objective of the autoencoder is to transport the data through a typically lower-dimensional latent space $\Z \subseteq \R^D$ while minimizing the reconstruction error. Thus, we search for an encoder $\E:X \to Z$ and decoder $\D: Z \to X$ functions, which minimize the reconstruction error between $X_i$ and its reconstructions $\D(\E X_i)$. We use a permutation invariant encoder based on PointNet architecture \cite{qi2017pointnet} and a modified decoder to produce weights instead of raw points. For point cloud representation, the crucial step is to define proper reconstruction loss that can be used in the auto-encoding framework. In our approach, we use NeRF cost function. 

\begin{figure}[!ht]
\includegraphics[width=0.9\linewidth]{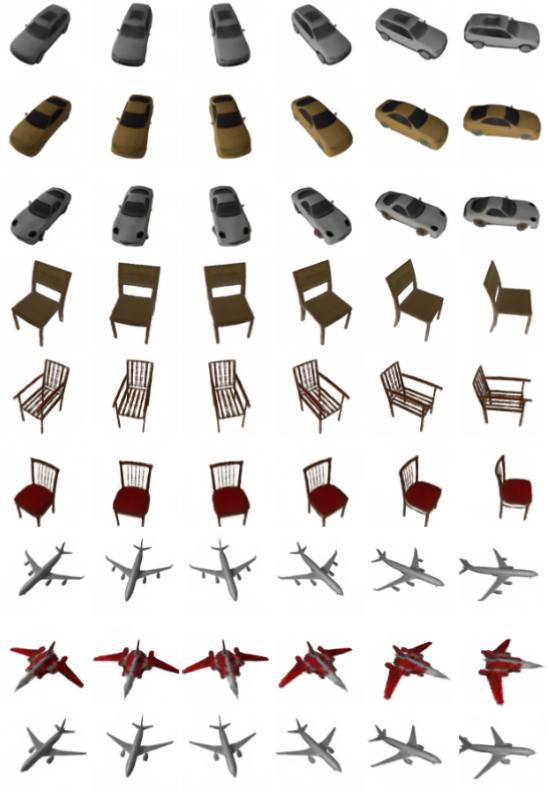}
\caption{Reconstructions of objects obtained by \our{}. }
\label{fig:rec_1} 
\end{figure}

\subsection{ \our{} }
%%%%%%%%%%%%%%%%%%%%%%%%%%%%%%%%%%%%

Our model uses three components: hypernetwork, autoencoder, and NeRF. The \our{} model uses Hypernetwork to output weights of generative network to create NeRF representation from the 3D point cloud. More specifically, we present parameterization of the 3D objects as a function $F_{\Theta} : \R^5 \to \R^4$, which given location $(x,y,z)$ and viewing direction $(\theta, \psi)$) returns a color ${\bf c} = (r, g, b)$ and volume density $\sigma$. 
Roughly speaking, instead of producing 3D objects, we would like to produce many neural networks (a different neural network for each object) that model them.

\begin{figure}[h]
\begin{center} 
    \includegraphics[width=0.90\linewidth]{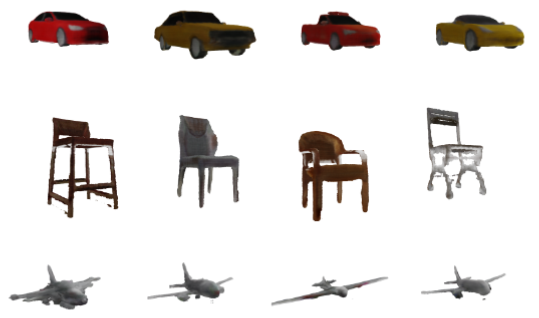} 
\end{center} 
\caption{From NeRF representation, we can extract 3D mesh with color. Mesh representation of objects produced by \our{}.}
\label{fig:rec_3D} 
\end{figure}

In practice, one neural network architecture uses different weights for each 3D object. The target network (NeRF) is not trained directly.   
We use a hypernetwork
$
\begin{array}{c}
H_{\Phi}: \R^3 \supset X \to \Theta ,
\end{array}
$
which for an point-cloud $X \subset \R^3$ returns weights $\Theta$ to the corresponding target network (NeRF) $F_{\Theta}$.
Thus, a point cloud $X$ is represented by a function:
$$
\begin{array}{c}
F((x,y,x,\theta,\psi);\Theta) = F((x,y,x,\theta,\psi); H_{\Phi}(X)).
\end{array}
$$

To use the above model, we need to train the weights $\Phi$ of the hypernetwork. For this purpose, we minimize the NeRF cost function over the training set consisting of pairs: point clouds and 2D images of the object. More precisely, we take an input point cloud $X \subset \R^6$ (the first three elements encode the position while the last three encode an RGB color) and pass it to $H_{\Phi}$. As a result, the hypernetwork returns weights $\Theta$ to the target network $F_{\Theta}$. Next, the set of 2D images is compared to the renderings generated by the target network $F_{\Theta}$. 
As a hypernetwork, we use a permutation invariant encoder based on PointNet architecture \cite{qi2017pointnet} and a modified decoder to produce weight instead of raw points.
The architecture of $H_{\Phi}$ consists of an encoder ($\E$) which is a PointNet-like network that transports the data to lower-dimensional latent space $\Z \in \R^D$ and a decoder ($\D$)  (fully-connected network), which transfers latent space to the vector of weights for the target network. In our framework, hypernetwork $H_{\Phi}(X)$ represents our autoencoder structure $\D(\E X)$.
Assuming  $H_{\Phi}(X) = \D(\E X)$, we train our model by minimizing the cost function given by equation (\ref{eq:cost_general}). 

We only train a single neural model (hypernetwork), which allows us to produce various functions at test time.
%%%%%%%%%%%%%%%%%%%%%%%%%%%%%%%%%%%%%%%%%%%%%

\begin{figure}[!ht]
\begin{center} 
\includegraphics[width=0.90\linewidth]{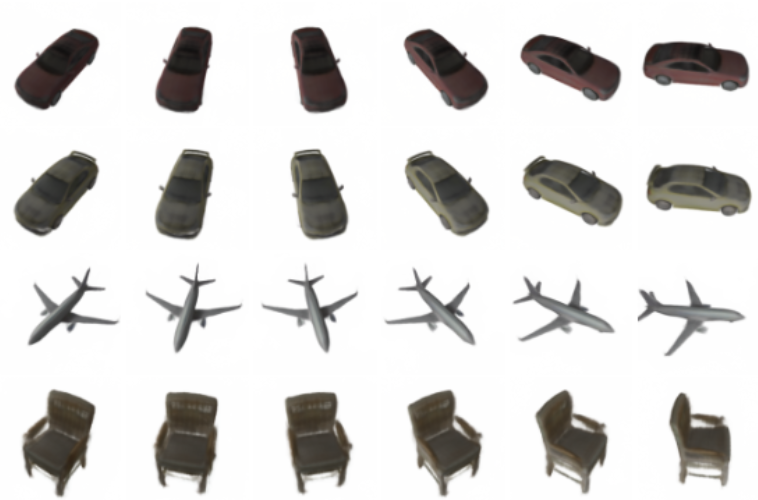} 
\end{center} 
\caption{Object generated by Generative \our{}.}
\label{fig:sample} 
\end{figure}

\subsection{Mesh representation}

In the paper, we represent 3D objects as Neural Radiance Fields. Such representation has few advantages over the classical one. In particular, we can obtain 3D mesh with colors. 

Thanks to volume density $\sigma$, we obtain voxel representation. We can predict an inside/outside category for points from grid $(x, y, z)$. Then we can render objects via the iso-surface extraction
a method such as Marching Cubes.
We can predict the color for all vertices when we have mesh representation. By using colors in the vertices of the mesh, we can add colors to the faces of the graph. We present 3D objects with colors in Fig.~\ref{fig:rec_3D}.

\begin{figure}[!ht]
\begin{center} 
\includegraphics[width=0.90\linewidth]{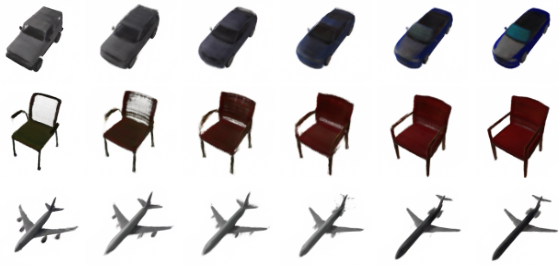} 
\end{center} 
\caption{Interpolations between elements produce by Generative \our{}.  }
\label{fig:interpolation} 
\end{figure}

%%%%%%%%%%%%%%%%%%%%%%%%%%%%%%%%%%%%
\subsection{Generative model}
%%%%%%%%%%%%%%%%%%%%%%%%%%%%%%%%%%%%

In our model, we use autoencoder architecture in hypernetwork. Therefore it is easy to construct a generative model.

\przemek{
An autoencoder-based generative model is a classical auto-encoder model with a modified cost function, which forces the model to be generative, i.e., ensures that the data transported to the latent space comes from the prior distribution (typically Gaussian)~\cite{kingma2013auto,tolstikhin2017wasserstein,knop2020cramer,knop2022generative}.
Thus, we use Variational Auto-encoders (VAE)~\cite{kingma2013auto} cost function to construct a generative auto-encoder model.
}

Variational Auto-encoders (VAE) are generative models capable of learning approximated data distribution by applying variational inference \cite{kingma2013auto}.
To ensure that the data transported to latent space $\Z$ are distributed according to standard normal density. We add the distance from standard multivariate normal density. By adding Kullback–Leibler divergence to our cost function to obtain a generative model, we cold Generative \our{}.
In Fig.~\ref{fig:sample}, we present samples, and in Fig.~\ref{fig:interpolation}, interpolations obtained by Generative \our{}.

% \begin{wraptable}{r}{0.52\textwidth}
\begin{table}
\begin{center}
    \begin{tabular}{ccc}
    & \our{} &  NeRF-KNN \\[0.1ex] 
     \hline
     planes (train) & \bf 24.83 &  13.75   \\[0.1ex] 
     planes (test) & \bf 20.45 & 14.18    \\[0.1ex] 
     cars (train) & \bf 28.14 &  11.82   \\[0.1ex]
     cars  (test) & \bf 20.86 & 12.47    \\[0.1ex]
     chairs (train) & \bf 23.90 &  10.02   \\[0.1ex]
     chairs  (test) & \bf 17.17 &  10.36    \\[0.1ex]
    \end{tabular}
\caption{Comparison of PSNR metric between our model and baseline KNN-NeRF trained on three classes of the ShapeNet data.}
% \przemek{, KNN rendering method on our cloud of points and classical NeRF trained on 8 different objects. Five random images per object from set were taken and mean was calculated. While NeRF used 25 test images.} }
\label{tab:1}
\end{center}
\vspace{-0.3cm}
\end{table}
% \end{wraptable}

%%%%%%%%%%%%%%%%%%%%%%%%%%%%%%%%%%%%%%%%%%%%%%%
\section{Experiments}
%%%%%%%%%%%%%%%%%%%%%%%%%%%%%%%%%%%%%%%%%%%%%%%

In this section, we describe the experimental results of the proposed model. To our knowledge, it is the first model that obtains translation from 3D point clouds to NeRF. \przemek{In practice, we use 3D point clouds as input, and other algorithms use 2D images. Furthermore, we use a generative model on many 3D objects, while most NeRF-based models are trained per one object. Therefore, it is hard to compare our results to other algorithms.} 
In the first subsection, we show that our model produces high-quality NeRF representations of the objects by comparing our model with our baseline KNN-NeRF. In the second one, we compare our model using a voxel representation obtained from NeRF. \przemek{At the end, we show that \our{} can be understood as a model for producing pre-train NeRF. }

% \begin{wraptable}{r}{0.52\textwidth}
% % \begin{table}
% \begin{center}
%     \begin{tabular}{c@{\;}c@{\;}c@{\;}c@{\;}c@{\;}} 
%     %  \hline
%       & \our{} & \our{} & NeRF-KNN & NeRF-KNN \\
%       & test set & train set & test set & train set   \\      
%       %[0.1ex] 
%      \hline
%      planes & 20.45 & 24.83 & 14.18 & 13.75   \\ 
%     %  \cline{1-3}
%      cars & 20.86 & 28.14 & 12.47 & 11.82   \\
%     %  \cline{1-3}
%      chairs & 17.17 & 23.90 & 10.36 & 10.02   \\
%     %  \hline
%     \end{tabular}
% \caption{Comparison of PSNR metric between our model and baseline KNN-NeRF trained on three classes of the ShapeNet data.}
% % \przemek{, KNN rendering method on our cloud of points and classical NeRF trained on 8 different objects. Five random images per object from set were taken and mean was calculated. While NeRF used 25 test images.} }
% \label{tab:1}
% \end{center}
% \vspace{-1cm}
% % \end{table}
% \end{wraptable}

%%%%%%%%%%%%%%%%%%%%%%%%%%%%%%%%%%%%%%%%%%%%%

\subsection{Methodology}

We used \textbf{ShapeNet} data set to train our model. First, we sampled $2048$ colored points from each object from three categories: cars, chairs, and planes. For each object: fifty $200$x$200$ transparent background images from random camera positions. 

% Data split was $80/20$.

\subsection{Point to NeRF evaluation}

We compare the metric reported by NeRF called PSNR (\textit{peak signal-to-noise ratio}) used to measure image reconstruction effectiveness. 

In Tab.~\ref{tab:1}, we compare \our{} and baseline solution KNN-NeRF. We present a comparison of the training and the test set. It should be highlighted that KNN-NeRF is trained separately for each object from the train and test set. The \our{} is trained on the training data set and evaluated on the test set.
\przemek{\our{} model achieves better results in all categories on the training and test sets. It confirms that training a single object using only point clouds is very hard.}

\subsection{Voxel representation}

NeRF produced by \our{} can produce 3D meshes. We can obtain mesh reconstruction for a given point cloud with the marching cubes algorithm.

To compare reconstruction with original mesh, we use Chamfer Distance,  defined as the distance between two clouds of points $P_1$ and $P_2$ such that:
\begin{equation}
    CD(P_1, P_2) \! = \! \frac{1}{2|P1|} \!\! \sum _{p_1 \in P1} \! \max_{p_2 \in P_2} d(p_1, p_2) + \frac{1}{2|P2|} \!\! \sum _{p_2 \in P2} \! \max_{p_1 \in P_1} d(p_1, p_2)
\end{equation}

Additionally, we use F-Score metric between two clouds of points with some threshold $t$ defined as:
\begin{equation}
\textrm{F-Score}(P_1, P_2, t) = \frac{2 \textrm{Recall Precision}}{\textrm{Recall + Precision}}.
\end{equation}
% where
% \begin{equation}
% \textrm{Recall}(P_1, P_2, t) = | \{ p_1 \in P_1, s.t. \max_{p_2 \in P_2} d(p_1, p_2) < t\}|
% \end{equation}

% \begin{equation}
% \textrm{Precision}(P_1, P_2, t) = | \{ p_2 \in P_2, s.t. \max_{p_1 \in P_1} d(p_2, p_1) < t\}|
% \end{equation}

For F-Score and Chamfer Distance calculation, we sample random $3000$ points from both original and reconstructed mesh, and we used threshold $t=0.01$ to find matching points for F-Score. 

\begin{table}
\begin{center}
{\small
    \begin{tabular}{c c c c c c} 
    %  \hline
     Category & PointConv & ONet & ConvONet & POCO & Ours \\ 
     \hline
     cars & 0.577 & 0.747 & 0.849 & 0.946 & \textbf{0.995}\\
    %  \hline
     planes & 0.562 & 0.829 & 0.965 & \textbf{0.994} & 0.952\\
    %  \hline
     chairs & 0.618 & 0.730 & 0.939 & 0.985 & \textbf{0.992}\\
    %  \hline
    \end{tabular}
}
\caption{F-Score comparison between our model and baselines: PointConv\cite{peng2020convolutional}, ONet\cite{mescheder2019occupancy}, ConvONet\cite{tang2021sa}, POCO\cite{poco}. We trained our model on $2048$ colored points and sampled on $3000$ while other methods used $3000$ to train and test. }
\label{tab:2}
\end{center}
\end{table}
% \vspace{-0.5cm}

\przemek{Even though our loss function was not directly related to mesh but to image reconstruction, we were able to achieve competitive results, see Tab~\ref{tab:2} and Tab~\ref{tab:3}.}

\begin{table}[h]
\begin{center}
{\small
    \begin{tabular}{c  c c c c c} 
    %  \hline
     Category & PointConv & ONet & ConvONet & POCO & Ours \\ 
     \hline
     \hline
     cars & 1.49 & 1.04 & 0.75 & 0.41 & \textbf{0.16}\\
    %  \hline
     planes & 1.40 & 0.64 & 0.34 & \textbf{0.23} & 0.91\\
    %  \hline
     chairs & 1.29 & 0.95 & 0.46 & 0.33 & \textbf{0.15}\\
    %  \hline
    \end{tabular}
    }
\caption{Chamfer Distance (multiplied by $10^2$) comparison between our model and baselines: PointConv, ONet, ConvONet, POCO. We trained our model on $2048$ colored points and sampled $3000$, while other methods used $3000$ to train and test.}
\label{tab:3}
\end{center}
\end{table}
\vspace{-0.5cm}

\begin{figure}[t!]
\begin{center}
\includegraphics[width=0.69\linewidth]{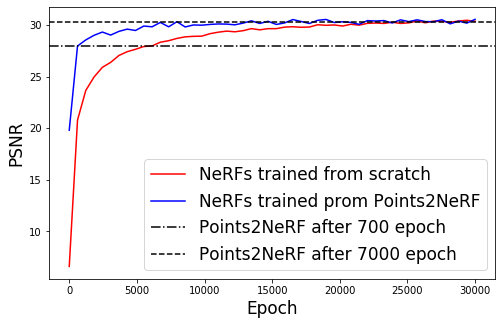}
\end{center}
\caption{ \our{} produced networks compared to the same NeRF networks trained from scratch. PSNR result is averaged on each object from the $20$ element subset of our test data set.  }
\label{fig:nerf_retrain} 
\end{figure}

\subsection{\our{} as a model for producing pre-training NeRF models} 

\przemek{
\our{} produce NeRF network from 3D point clouds. The quality of NeRF model train on 3D point clouds is slightly worse than the classical image based. On the other hand, when \our{} is trained, we can produce NeRF very effectively. Therefore, we can reduce time and resources for training classical image-based NeRF by using weights produced by \our{} as an initialization. 
}

\przemek{
To compare the time-saving efficiency of produced networks with traditionally trained NeRFs, we performed an additional experiment where we trained $20$ NeRFs using random objects of our test data set. Additionally, we used weights generated by \our{} method as our starting point.
}

\przemek{
In Fig.~\ref{fig:nerf_retrain}, we present PSNR obtained by classical NeRF and NeRF initialized by \our{}.
As we can see, our initialization allows us to obtain a good level of PSNR in approximately 7000 epochs, and classical NeRF needs 30 000 epochs.
}

\subsection{Hardware Requirements}
\przemek{The large hypernetwork architecture of \our{} needs substantial time and computational resources to be optimized. We trained our model on the ShapeNet car category data set containing $2797$ objects for $2200$ epochs on Tesla V100-SXM2 GPU, which in total required 300 hours of calculation. }

\section{Conclusions}
\label{sec:conclusions}

In this work, we presented a novel approach to generating NeRF representation from 3D point clouds. Our model leverages a hypernetwork paradigm and NeRF representation of the 3D scene. \our{} take a 3D point cloud with the associated color values and return the weights of a NeRF network that reconstructs 3D objects from 2D images. Such representation gives several advantages over the existing approaches. First, we can add a conditioning mechanism to NeRFs that allows the creation of a generative model. Secondly, we can quickly obtain mesh representation with colors, which is a challenging task in 3D object rendering.

In future work, we plan to modify the architecture to work on the partial and noisy point clouds. 

\section{Acknowledgements}

The research of D. Zimny was supported by the National Centre of Science (Poland) Grant No. 2021/43/B/ST6/01456.
The research of T. Trzcinski was carried out within the research project "Bio-inspired artificial neural network" (grant no. POIR.04.04.00-00-14DE/18-00) within the Team-Net program of the Foundation for Polish Science co-financed by the European Union under the European Regional Development Fund. It was partially funded by National Science Centre, Poland (grant no 2020/39/B/ST6/01511 and 2022/45/B/ST6/02817).
The research of P. Spurek has been supported by a flagship project entitled ``Artificial Intelligence Computing Center Core Facility'' from the DigiWorld Priority Research Area under the Strategic Programme Excellence Initiative at Jagiellonian University.

\bibliographystyle{plain}
%\bibliography{references}

\begin{thebibliography}{46}
\providecommand{\natexlab}[1]{#1}
\providecommand{\url}[1]{\texttt{#1}}
\expandafter\ifx\csname urlstyle\endcsname\relax
  \providecommand{\doi}[1]{doi: #1}\else
  \providecommand{\doi}{doi: \begingroup \urlstyle{rm}\Url}\fi

\bibitem[Achlioptas et~al.(2018)Achlioptas, Diamanti, Mitliagkas, and
  Guibas]{achlioptas2018learning}
P.~Achlioptas, O.~Diamanti, I.~Mitliagkas, and L.~Guibas.
\newblock Learning representations and generative models for 3d point clouds.
\newblock In \emph{International conference on machine learning}, pages 40--49.
  PMLR, 2018.

\bibitem[Arsalan~Soltani et~al.(2017)Arsalan~Soltani, Huang, Wu, Kulkarni, and
  Tenenbaum]{arsalan2017synthesizing}
A.~Arsalan~Soltani, H.~Huang, J.~Wu, T.~D. Kulkarni, and J.~B. Tenenbaum.
\newblock Synthesizing 3d shapes via modeling multi-view depth maps and
  silhouettes with deep generative networks.
\newblock In \emph{Proceedings of the IEEE conference on computer vision and
  pattern recognition}, pages 1511--1519, 2017.

\bibitem[Barron et~al.(2021)Barron, Mildenhall, Tancik, Hedman, Martin-Brualla,
  and Srinivasan]{barron2021mip}
J.~T. Barron, B.~Mildenhall, M.~Tancik, P.~Hedman, R.~Martin-Brualla, and P.~P.
  Srinivasan.
\newblock Mip-nerf: A multiscale representation for anti-aliasing neural
  radiance fields.
\newblock In \emph{Proceedings of the IEEE/CVF International Conference on
  Computer Vision}, pages 5855--5864, 2021.

\bibitem[Boulch and Marlet(2022)]{poco}
A.~Boulch and R.~Marlet.
\newblock Poco: Point convolution for surface reconstruction.
\newblock In \emph{Proceedings of the IEEE/CVF Conference on Computer Vision
  and Pattern Recognition}, pages 6302--6314, 2022.

\bibitem[Cai et~al.(2020)Cai, Yang, Averbuch-Elor, Hao, Belongie, Snavely, and
  Hariharan]{cai2020learning}
R.~Cai, G.~Yang, H.~Averbuch-Elor, Z.~Hao, S.~Belongie, N.~Snavely, and
  B.~Hariharan.
\newblock Learning gradient fields for shape generation.
\newblock In \emph{Computer Vision--ECCV 2020: 16th European Conference,
  Glasgow, UK, August 23--28, 2020, Proceedings, Part III 16}, pages 364--381.
  Springer, 2020.

\bibitem[Chen and Zhang(2019)]{chen2019learning}
Z.~Chen and H.~Zhang.
\newblock Learning implicit fields for generative shape modeling.
\newblock In \emph{Proceedings of the IEEE/CVF Conference on Computer Vision
  and Pattern Recognition}, pages 5939--5948, 2019.

\bibitem[Choy et~al.(2016)Choy, Xu, Gwak, Chen, and Savarese]{choy20163d}
C.~B. Choy, D.~Xu, J.~Gwak, K.~Chen, and S.~Savarese.
\newblock 3d-r2n2: A unified approach for single and multi-view 3d object
  reconstruction.
\newblock In \emph{European conference on computer vision}, pages 628--644.
  Springer, 2016.

\bibitem[Dupont et~al.(2022)Dupont, Teh, and Doucet]{dupont2021generative}
E.~Dupont, Y.~W. Teh, and A.~Doucet.
\newblock Generative models as distributions of functions.
\newblock In \emph{International Conference on Artificial Intelligence and
  Statistics}, pages 2989--3015. PMLR, 2022.

\bibitem[Fridovich-Keil et~al.(2022)Fridovich-Keil, Yu, Tancik, Chen, Recht,
  and Kanazawa]{fridovich2022plenoxels}
S.~Fridovich-Keil, A.~Yu, M.~Tancik, Q.~Chen, B.~Recht, and A.~Kanazawa.
\newblock Plenoxels: Radiance fields without neural networks.
\newblock In \emph{Proceedings of the IEEE/CVF Conference on Computer Vision
  and Pattern Recognition}, pages 5501--5510, 2022.

\bibitem[Girdhar et~al.(2016)Girdhar, Fouhey, Rodriguez, and
  Gupta]{girdhar2016learning}
R.~Girdhar, D.~F. Fouhey, M.~Rodriguez, and A.~Gupta.
\newblock Learning a predictable and generative vector representation for
  objects.
\newblock In \emph{European Conference on Computer Vision}, pages 484--499.
  Springer, 2016.

\bibitem[Grathwohl et~al.(2018)Grathwohl, Chen, Bettencourt, Sutskever, and
  Duvenaud]{grathwohl2018ffjord}
W.~Grathwohl, R.~T. Chen, J.~Bettencourt, I.~Sutskever, and D.~Duvenaud.
\newblock Ffjord: Free-form continuous dynamics for scalable reversible
  generative models.
\newblock In \emph{International Conference on Learning Representations}, 2018.

\bibitem[Ha et~al.(2017)Ha, Dai, and Le]{ha2016hypernetworks}
D.~Ha, A.~Dai, and Q.~V. Le.
\newblock Hypernetworks.
\newblock In \emph{International Conference on Learning Representations}, 2017.

\bibitem[H{\"a}ne et~al.(2017)H{\"a}ne, Tulsiani, and
  Malik]{hane2017hierarchical}
C.~H{\"a}ne, S.~Tulsiani, and J.~Malik.
\newblock Hierarchical surface prediction for 3d object reconstruction.
\newblock In \emph{2017 International Conference on 3D Vision (3DV)}, pages
  412--420. IEEE, 2017.

\bibitem[Kajiya and Von~Herzen(1984)]{kajiya1984ray}
J.~T. Kajiya and B.~P. Von~Herzen.
\newblock Ray tracing volume densities.
\newblock \emph{ACM SIGGRAPH computer graphics}, 18\penalty0 (3):\penalty0
  165--174, 1984.

\bibitem[Kania et~al.(2022)Kania, Yi, Kowalski, Trzci{\'n}ski, and
  Tagliasacchi]{kania2021conerf}
K.~Kania, K.~M. Yi, M.~Kowalski, T.~Trzci{\'n}ski, and A.~Tagliasacchi.
\newblock Conerf: Controllable neural radiance fields.
\newblock In \emph{Proceedings of the IEEE/CVF Conference on Computer Vision
  and Pattern Recognition}, pages 18623--18632, 2022.

\bibitem[Kingma and Welling(2014)]{kingma2013auto}
D.~P. Kingma and M.~Welling.
\newblock Auto-encoding variational bayes.
\newblock In \emph{International Conference on Learning Representations (ICLR
  2014)}. OpenReview.net, 2014.

\bibitem[Knop et~al.(2020)Knop, Spurek, Tabor, Podolak, Mazur, and
  Jastrz{\k{e}}bski]{knop2020cramer}
S.~Knop, P.~Spurek, J.~Tabor, I.~Podolak, M.~Mazur, and S.~Jastrz{\k{e}}bski.
\newblock Cramer-wold auto-encoder.
\newblock \emph{Journal of Machine Learning Research}, 21, 2020.

\bibitem[Knop et~al.(2022)Knop, Mazur, Spurek, Tabor, and
  Podolak]{knop2022generative}
S.~Knop, M.~Mazur, P.~Spurek, J.~Tabor, and I.~Podolak.
\newblock Generative models with kernel distance in data space.
\newblock \emph{Neurocomputing}, 487:\penalty0 119--129, 2022.

\bibitem[Kosiorek et~al.(2021)Kosiorek, Strathmann, Zoran, Moreno, Schneider,
  Mokr{\'a}, and Rezende]{kosiorek2021nerf}
A.~R. Kosiorek, H.~Strathmann, D.~Zoran, P.~Moreno, R.~Schneider, S.~Mokr{\'a},
  and D.~J. Rezende.
\newblock Nerf-vae: A geometry aware 3d scene generative model.
\newblock In \emph{International Conference on Machine Learning}, pages
  5742--5752. PMLR, 2021.

\bibitem[Li et~al.(2017)Li, Xu, Chaudhuri, Yumer, Zhang, and
  Guibas]{li2017grass}
J.~Li, K.~Xu, S.~Chaudhuri, E.~Yumer, H.~Zhang, and L.~Guibas.
\newblock Grass: Generative recursive autoencoders for shape structures.
\newblock \emph{ACM Transactions on Graphics (TOG)}, 36\penalty0 (4):\penalty0
  1--14, 2017.

\bibitem[Liu et~al.(2020)Liu, Gu, Zaw~Lin, Chua, and Theobalt]{liu2020neural}
L.~Liu, J.~Gu, K.~Zaw~Lin, T.-S. Chua, and C.~Theobalt.
\newblock Neural sparse voxel fields.
\newblock \emph{Advances in Neural Information Processing Systems},
  33:\penalty0 15651--15663, 2020.

\bibitem[Liu et~al.(2022)Liu, Zhang, Gao, and Wang]{LIU2022108774}
Z.~Liu, Y.~Zhang, J.~Gao, and S.~Wang.
\newblock Vfmvac: View-filtering-based multi-view aggregating convolution for
  3d shape recognition and retrieval.
\newblock \emph{Pattern Recognition}, 129:\penalty0 108774, 2022.
\newblock ISSN 0031-3203.

\bibitem[Max(1995)]{max1995optical}
N.~Max.
\newblock Optical models for direct volume rendering.
\newblock \emph{IEEE Transactions on Visualization and Computer Graphics},
  1\penalty0 (2):\penalty0 99--108, 1995.

\bibitem[Mescheder et~al.(2019)Mescheder, Oechsle, Niemeyer, Nowozin, and
  Geiger]{mescheder2019occupancy}
L.~Mescheder, M.~Oechsle, M.~Niemeyer, S.~Nowozin, and A.~Geiger.
\newblock Occupancy networks: Learning 3d reconstruction in function space.
\newblock In \emph{Proceedings of the IEEE/CVF Conference on Computer Vision
  and Pattern Recognition}, pages 4460--4470, 2019.

\bibitem[Michalkiewicz et~al.(2019)Michalkiewicz, Pontes, Jack, Baktashmotlagh,
  and Eriksson]{michalkiewicz2019implicit}
M.~Michalkiewicz, J.~K. Pontes, D.~Jack, M.~Baktashmotlagh, and A.~Eriksson.
\newblock Implicit surface representations as layers in neural networks.
\newblock In \emph{Proceedings of the IEEE/CVF International Conference on
  Computer Vision}, pages 4743--4752, 2019.

\bibitem[Mildenhall et~al.(2020)Mildenhall, Srinivasan, Tancik, Barron,
  Ramamoorthi, and Ng]{mildenhall2020nerf}
B.~Mildenhall, P.~P. Srinivasan, M.~Tancik, J.~T. Barron, R.~Ramamoorthi, and
  R.~Ng.
\newblock Nerf: Representing scenes as neural radiance fields for view
  synthesis.
\newblock In \emph{European conference on computer vision}, pages 405--421.
  Springer, 2020.

\bibitem[Niemeyer et~al.(2020)Niemeyer, Mescheder, Oechsle, and
  Geiger]{niemeyer2020differentiable}
M.~Niemeyer, L.~Mescheder, M.~Oechsle, and A.~Geiger.
\newblock Differentiable volumetric rendering: Learning implicit 3d
  representations without 3d supervision.
\newblock In \emph{Proceedings of the IEEE/CVF Conference on Computer Vision
  and Pattern Recognition}, pages 3504--3515, 2020.

\bibitem[Ost et~al.(2022)Ost, Laradji, Newell, Bahat, and Heide]{ost2021neural}
J.~Ost, I.~Laradji, A.~Newell, Y.~Bahat, and F.~Heide.
\newblock Neural point light fields.
\newblock In \emph{Proceedings of the IEEE/CVF Conference on Computer Vision
  and Pattern Recognition}, pages 18419--18429, 2022.

\bibitem[Park et~al.(2019)Park, Florence, Straub, Newcombe, and
  Lovegrove]{park2019deepsdf}
J.~J. Park, P.~Florence, J.~Straub, R.~Newcombe, and S.~Lovegrove.
\newblock Deepsdf: Learning continuous signed distance functions for shape
  representation.
\newblock In \emph{Proceedings of the IEEE/CVF Conference on Computer Vision
  and Pattern Recognition}, pages 165--174, 2019.

\bibitem[Peng et~al.(2020)Peng, Niemeyer, Mescheder, Pollefeys, and
  Geiger]{peng2020convolutional}
S.~Peng, M.~Niemeyer, L.~Mescheder, M.~Pollefeys, and A.~Geiger.
\newblock Convolutional occupancy networks.
\newblock In \emph{Computer Vision--ECCV 2020: 16th European Conference,
  Glasgow, UK, August 23--28, 2020, Proceedings, Part III 16}, pages 523--540.
  Springer, 2020.

\bibitem[Qi et~al.(2017)Qi, Su, Mo, and Guibas]{qi2017pointnet}
C.~R. Qi, H.~Su, K.~Mo, and L.~J. Guibas.
\newblock Pointnet: Deep learning on point sets for 3d classification and
  segmentation.
\newblock In \emph{Proceedings of the IEEE Conference on Computer Vision and
  Pattern Recognition}, pages 652--660, 2017.

\bibitem[Rahaman et~al.(2019)Rahaman, Baratin, Arpit, Draxler, Lin, Hamprecht,
  Bengio, and Courville]{rahaman2019spectral}
N.~Rahaman, A.~Baratin, D.~Arpit, F.~Draxler, M.~Lin, F.~Hamprecht, Y.~Bengio,
  and A.~Courville.
\newblock On the spectral bias of neural networks.
\newblock In \emph{International Conference on Machine Learning}, pages
  5301--5310. PMLR, 2019.

\bibitem[Reiser et~al.(2021)Reiser, Peng, Liao, and Geiger]{reiser2021kilonerf}
C.~Reiser, S.~Peng, Y.~Liao, and A.~Geiger.
\newblock Kilonerf: Speeding up neural radiance fields with thousands of tiny
  mlps.
\newblock In \emph{Proceedings of the IEEE/CVF International Conference on
  Computer Vision}, pages 14335--14345, 2021.

\bibitem[Sheikh et~al.(2017)Sheikh, Rasul, Merentitis, and
  Bergmann]{sheikh2017stochastic}
A.-S. Sheikh, K.~Rasul, A.~Merentitis, and U.~Bergmann.
\newblock Stochastic maximum likelihood optimization via hypernetworks.
\newblock \emph{arXiv preprint arXiv:1712.01141}, 2017.

\bibitem[Shu et~al.(2022)Shu, Park, and Kwon]{shu2022wasserstein}
D.~W. Shu, S.~W. Park, and J.~Kwon.
\newblock Wasserstein distributional harvesting for highly dense 3d point
  clouds.
\newblock \emph{Pattern Recognition}, 132:\penalty0 108978, 2022.

\bibitem[Sinha et~al.(2016)Sinha, Bai, and Ramani]{sinha2016deep}
A.~Sinha, J.~Bai, and K.~Ramani.
\newblock Deep learning 3d shape surfaces using geometry images.
\newblock In \emph{European Conference on Computer Vision}, pages 223--240.
  Springer, 2016.

\bibitem[Sitzmann et~al.(2019)Sitzmann, Zollh{\"o}fer, and
  Wetzstein]{sitzmann2019scene}
V.~Sitzmann, M.~Zollh{\"o}fer, and G.~Wetzstein.
\newblock Scene representation networks: Continuous 3d-structure-aware neural
  scene representations.
\newblock \emph{Advances in Neural Information Processing Systems}, 32, 2019.

\bibitem[Spurek et~al.(2020)Spurek, Winczowski, Tabor, Zamorski, Zieba, and
  Trzcinski]{spurek2020hypernetwork}
P.~Spurek, S.~Winczowski, J.~Tabor, M.~Zamorski, M.~Zieba, and T.~Trzcinski.
\newblock Hypernetwork approach to generating point clouds.
\newblock In \emph{International Conference on Machine Learning}, pages
  9099--9108. PMLR, 2020.

\bibitem[Spurek et~al.(2021)Spurek, Zieba, Tabor, and
  Trzcinski]{spurek2021general}
P.~Spurek, M.~Zieba, J.~Tabor, and T.~Trzcinski.
\newblock General hypernetwork framework for creating 3d point clouds.
\newblock \emph{IEEE Transactions on Pattern Analysis and Machine
  Intelligence}, 2021.

\bibitem[Spurek et~al.(2022)Spurek, Kasymov, Mazur, Janik, Tadeja, Struski,
  Tabor, and Trzci{\'n}ski]{spurek2021hyperpocket}
P.~Spurek, A.~Kasymov, M.~Mazur, D.~Janik, S.~Tadeja, L.~Struski, J.~Tabor, and
  T.~Trzci{\'n}ski.
\newblock Hyperpocket: Generative point cloud completion.
\newblock In \emph{2022 IEEE/RSJ International Conference on Intelligent Robots
  and Systems (IROS)}, 2022.

\bibitem[Tang et~al.(2021)Tang, Lei, Xu, Ma, Jia, and Zhang]{tang2021sa}
J.~Tang, J.~Lei, D.~Xu, F.~Ma, K.~Jia, and L.~Zhang.
\newblock Sa-convonet: Sign-agnostic optimization of convolutional occupancy
  networks.
\newblock In \emph{Proceedings of the IEEE/CVF International Conference on
  Computer Vision}, pages 6504--6513, 2021.

\bibitem[Tolstikhin et~al.(2018)Tolstikhin, Bousquet, Gelly, and
  Sch{\"o}lkopf]{tolstikhin2017wasserstein}
I.~Tolstikhin, O.~Bousquet, S.~Gelly, and B.~Sch{\"o}lkopf.
\newblock Wasserstein auto-encoders.
\newblock In \emph{6th International Conference on Learning Representations
  (ICLR 2018)}. OpenReview.net, 2018.

\bibitem[Wang et~al.(2021)Wang, Liu, Liu, Theobalt, Komura, and
  Wang]{wang2021neus}
P.~Wang, L.~Liu, Y.~Liu, C.~Theobalt, T.~Komura, and W.~Wang.
\newblock Neus: Learning neural implicit surfaces by volume rendering for
  multi-view reconstruction.
\newblock \emph{Advances in Neural Information Processing Systems},
  34:\penalty0 27171--27183, 2021.

\bibitem[Xu et~al.(2022)Xu, Xu, Philip, Bi, Shu, Sunkavalli, and
  Neumann]{xu2022point}
Q.~Xu, Z.~Xu, J.~Philip, S.~Bi, Z.~Shu, K.~Sunkavalli, and U.~Neumann.
\newblock Point-nerf: Point-based neural radiance fields.
\newblock In \emph{Proceedings of the IEEE/CVF Conference on Computer Vision
  and Pattern Recognition}, pages 5438--5448, 2022.

\bibitem[Yang et~al.(2022)Yang, Davoine, Wang, and Jin]{yang2022continuous}
F.~Yang, F.~Davoine, H.~Wang, and Z.~Jin.
\newblock Continuous conditional random field convolution for point cloud
  segmentation.
\newblock \emph{Pattern Recognition}, 122:\penalty0 108357, 2022.

\bibitem[Yang et~al.(2019)Yang, Huang, Hao, Liu, Belongie, and
  Hariharan]{yang2019pointflow}
G.~Yang, X.~Huang, Z.~Hao, M.-Y. Liu, S.~Belongie, and B.~Hariharan.
\newblock Pointflow: 3d point cloud generation with continuous normalizing
  flows.
\newblock In \emph{Proceedings of the IEEE International Conference on Computer
  Vision}, pages 4541--4550, 2019.

\end{thebibliography}

\end{document}